# NARRATIVE SCIENCE SYSTEMS: A REVIEW

[1]Paramjot Kaur Sarao, [2]Puneet Mittal, [3]Rupinder Kaur
[1]M.Tech (IT), Baba Banda Singh Bahadur Engineering College, Fatehgarh Sahib, Punjab, INDIA
Email: param.sarao@gmail.com
[2]Assistant Professor, CSE/IT Department, Baba Banda Singh Bahadur Engineering College, Fatehgarh Sahib
Punjab, INDIA
Email: puneet.mittal@bbsbec.ac.in
[3]M.Tech (IT), Chandigarh Engineering College, Landran, Punjab, INDIA
Email: rupinderksarao1992@gmail.com

***Abstract:*** *Automatic narration of events and entities is the need of the hour, especially when live reporting is critical and volume of information to be narrated is huge. This paper discusses the challenges in this context, along with the algorithms used to build such systems. From a systematic study, we can infer that most of the work done in this area is related to statistical data. It was also found that subjective evaluation or contribution of experts is also limited for narration context.*

***Keywords:*** *computational narration, automatic story generation, automatic narration, content creation, natural language processing.*

## I. INTRODUCTION

Today, large volume of data is curated, sorted and stored in different architectures, simply in a hope that some meaningful insight, some discoverable pattern or knowledge may appear either in contemporary context of problems or for future applications.

Based on these data collections, many researchers are questioning the fact, what it takes to build a system that can narrate a story by reading the data, suggest, advice humans on their problems that can be identified from the data patterns and their semantic representations [5]. However, before researchers find the solution, the first problem is to set criteria to capture the 'category' of the objects to be 'suggested', 'recommended'[1] or narrated for problems.

An intelligent configurable [2] 'software', algorithm cannot treat every object it sees as unique entity unlike anything else in universe. It has to put objects in 'category' [12] that it may apply its hard-won knowledge about similar objects, encountered in the past, to the object at hand.

Any robotic device's brain must first understand by using machine learning algorithm the cultural, semantic context of the 'category'[3] of the problem and its possible solutions for narrating its story to humans, nuances that would come as narration/recommendations /suggestions, which go beyond predicting the sequence of events, formats, plots of story/block of content [2] etc.

Is the narration process[7], a process of finding similar objects for story telling[4] only or it also understands those objects, their anatomy, their structure to recommend solutions and insights, especially in the case of technical suggestions? Is it possible to open up concepts with some algorithmic soft computing screws to drives and read its 'make' or simply reverse engineer it and suggest a better solution, that too a technical solution one to narrate a human understandable story along with ranking of solution quality [12] ?

Is there a possibility that a robotic algorithm conducts a crowd sourcing operation in the background using cloud [19] for suggesting solutions in real time and narrate story in human like voice from sensor data [14]?

The universe not only obeys the laws of nature fundamentally but it also takes shape based on series of accidents or actions [18] to reach where it is now. Is it possible that a recommendation or narrative system undergoes a series of 'accidents' to discover new knowledge, processes, laws and even hypothesis to obtain some conclusive 'advice to tell humans' for some issue or a challenge?

The current narrative systems, especially in the area of digitalized platform or electronic data [13] observe the 'data patterns' of the stored data or the sequence of the objects written in tabular, mathematical sense. Typically when, one faces some issue, challenge or needs to complete some task, the first help comes essentially from our own human brain. Any mathematical model that mimics our brain activity, can suggest solutions, recommends things from 'experience', as well as from knowledge/data stored/from event streams [20] for narrating a story out of it. But, at the same time we must be aware that human brains are slow whereas computers are fast, human brains are noisy and computers are not. Human brains have trillions processing units called neurons to process information whereas computers to do not have process at this scale. Even if we go to largest scale of deployment of distributed computers, both human brains and computers may have partial information or non – observable information which forces us to think in terms of algorithms that can raise the levels of accuracy in narrating the 'story items'. Therefore, the next section of this paper systematically explores algorithms, methods used for building narration systems.

## II. RELATED WORK

Allen [1] has presented his research on the phrase variation of a template, essentially trying to build dynamic templates by adding variations to its phrases, sentences. This was done by building an associative matrix of each phrase in the template with its possible permutation, combination and variations. In simple words, two-three phrases, idioms may have similar meaning, context and even similar semantics. Based on this idea the author has been able to build a system that can generate narrative content with lot of variations due to dynamic selection of phrases with respect to narrative context of some real world event.

Birnbaum et al. [4] described a narrative content generation system that received data and information pertaining to a domain event like sports match. The received data and information and one or more derived features were then used to identify the variation of angles for a narrative story. These angles / aspects were further filtered by the parameters that specify the length of the story, focus of the story etc. The points associated with the filtered plurality of the angles were then aggregated for final story generation output.

Schilder [6] has claimed regarding a narrative system that work on data feed which has been provided by a statistical provider. A statistical provider is an entity defined as a resource that provides raw and processed data in structured form with mathematical tabulation, calculations and summaries. The inventor here subjects this mathematical embodiment to pre-segmented text for narration generation. However, theses narrations were produced finally based on a clustered plurals (category) text, in conjunction with templates to produce natural language text.

Nichols et al. [7] presented a work based on artificial intelligence methods. The patent applicants assumed that each story to be narrated have multiple aspects/angles which creates plurality in story. However, this plurality must be evaluated or simply must be given some rank, category etc., so that its selection was ordered for find narration content generation. The system angle data set structure which gave dynamic variations, a subject data components which helped in selection or domain for which the story was being developed for. Then, they have a story evaluator component, which gave evaluation indicators for selection or was put in story generation queue to produce a narrative story. The source data about subject and angles undergo variations, testing and applicability test for story under consideration and then finally derived features were computed for story generation.

Birnbaum et al. [8] presented a research paper related to generating narrative stories automatically from data that may be gathered from diverse resources like web scrapping. Once data was sourced, its features were derived in conjunction with perspectives/ aspects/ angles of the story. A priority list was built for these angles using deep filtering to finally generate the narrative output.

Riedl et al. [9] have tried to relate story plots, formats, and templates with artificial intelligence. Typically, journalists, novelists, writers may write real life stories that would essentially have some scheme of things, style, format, pattern based on which they author or narrator to the audiences. This research paper explored the possibilities of building similar capabilities using computer algorithms. The paper helped in understanding narrative planning methods.

Ghuman et al. [10] have developed an algorithm that works in cricket domain. The system feeds on statistical information of the cricket tournaments. The algorithm built a decision tree based on the information bytes contained in a phrase which described some cricketing act. For calculating information gained entropy measurements were done which finally helped to select the order of narrative event rendering which described the cricket match or tournament.

Novalija et al. [11] presented an automatic narration system that also worked on the statistical source(s) which were restricted to particular domain. The authors had worked in web analytics domain and received data from Google Analytics module to build a decision tree. The authors have also addressed multiple reporting scenarios by using multiple rules that build templates on the fly. Last but not the least they used machine learning algorithm called C4.5 for classification of the statistical data.

*Table 1: Analysis of previous work on the basis of different aspects:*

| Characteristics ⟶ | Data type Source | Domain | Algorithm | Filters | Narration Output Quality Check | Narration Quality Truth Validation | Human Subject Agreement On Phrases |
|---|---|---|---|---|---|---|---|
| Allen [1] | Statistical | Event Based | Variation of Phrases | Multiple Rules | No | Yes | No |
| Birnbaum et al. [4] | Statistical | Sports | Order setting | Narration Length | No | Yes | No |
| Schilder [6] | Statistical | NA | Clustering | Narration Length | NA | No | No |
| Nichols et al. [7] | Statistical | Sports | Variation of Angles | Narration Length | No | No | No |
| Birnbaum et al. [8] | Web Scrapping Content | Any | Priority ordering | Multiple Rules | No | No | No |
| Riedl et al. [9] | Generic Human Acts Text | Human Acts | Intent Based partial Order Casual Link | Multiple Rules | No | Yes | No |
| Ghuman et al. [10] | Statistical | Cricket | Entropy Based | Narration Length | No | No | No |
| Novalija et al. [11] | Statistical | Web Analytics | Decision Tree | Narration Length | No | No | No |

It is apparent from the tabular analysis that most of the narration systems work on statistical formatted data and very few systems are working on text /content data. And major emphasis of research in this area is on narration of sports and events related to statistical information. The narrations are normally rule based and most important rule is 'length' of the narration. Extensive text processing algorithms are required which include clustering algorithms [14], variation of phrases, variation of angles, Intent Based Partial Order Casual Link etc. It is clear from the above Table 1 that, patents and research papers solicited in context of automatic narration generation have ignored work related to validation and Quality check of narration output by human subject and domain experts.

## III. ARCHITECTURE OF NARRATION GENERATION SYSTEM

The block diagram (figure 1) shows, the typical components and process that are used for building automatic narration system.

1. The System has five main modules. The first module consists of Database of Domain specific taxonomy, ontology, phrases, sentences and event descriptors for focused on particular domain like sports.

2. The second component consists of statistical data related to particular domain like sports. This database may be RDBMS [21] or Distributed Storage like HDFS [22].

3. Each of these statistical Data, phrases must be collaborated with particular event format or plot. Hence each narration structure is defined in templates /or formats / or blueprint sets [2]. From this format, the automatic narration generation algorithm picks its structure for creating 'output' narration.

4. The output narration generation algorithm may be based on clustering, Artificial Intelligence [23], variation of angles, order setting, Priority ordering [8], Entropy Based and Decision tree [24].

5. The Fifth module is the automated narration output based on above said algorithms. The output may be html, pdf, word document, audio [17], video narration [16] or Mixed media [15].

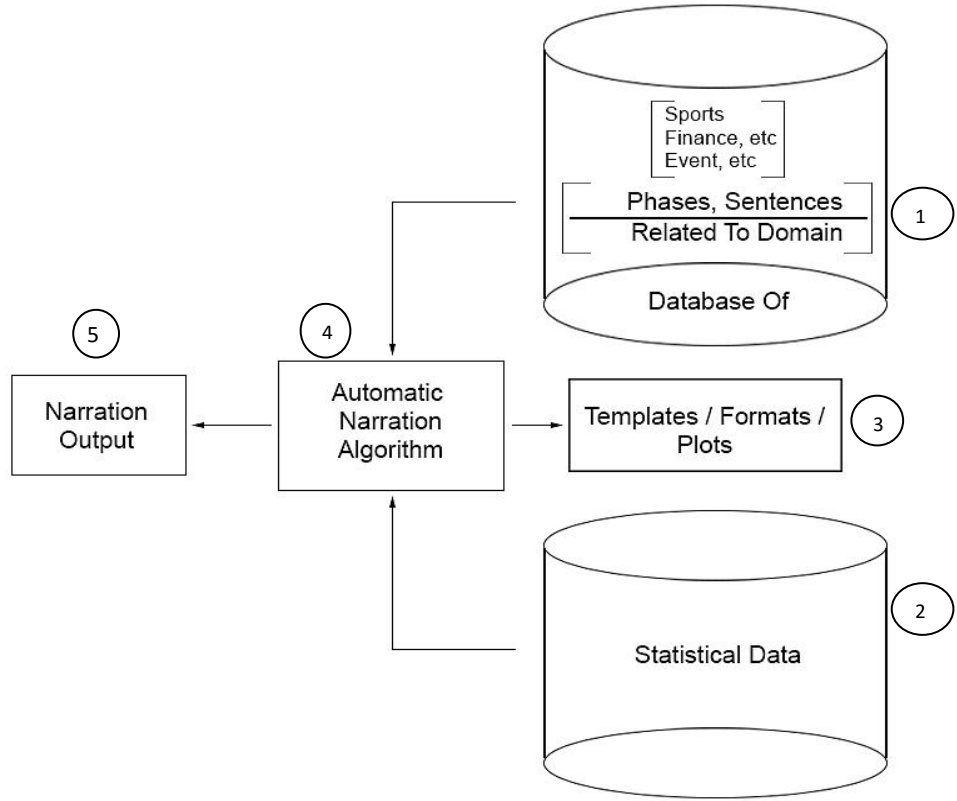

## IV. DISCUSSION

Major work done in this context, are essentially suggesting story based on 10-15 entities or objects which can be structured. The structure of such context will depend upon the domain e.g. sports and the entities/objects it can hold to describe the event tastefully with high quality. However, it was also found that there is trust deficit in context of the 'content quality' of automatically generated narration, as builders of such systems are not engaging much human subjects to evaluate quality of automatically generated narration, nor the ground truth of such content is checked aggressively by multiple reviewers, using methods like blind peer-review or by using methods like Delphi. Many systems found in literature in this area are based on Artificial Intelligence and probability based algorithms which offer various degrees of accuracy and efficiency and are limited mostly to descriptive and mathematical explanation of the events.

## V. CONCLUSION

In summary, we can infer from the above study that there is ample scope to work with new inputs for improving automatic narration generation system. Improvement not just in the methods in collaboration of events to plots/formats/structure of narration but also in context of using 'online collaboration' or using 'crowd wisdom' to select high quality of "narration". The existing solutions and algorithms used to build narration are already authoring narratives for businesses scenarios where data is a tool to build products and organizations. The existing algorithms works on principles of probability (Decision Trees, Bayes) and Artificial intelligence (Neural networks), which would remain the main components of this narrative science algorithms. However, within the realm of these deep learning algorithms with human based subjective evaluation system is the future, since," Combinational" language learning has  limited scope in understanding the semantic and cultural context of the narration especially when we try to build systems that are not based on structure / format / plot. Second area which needs to be emphasized is Visualization of the data along with narration and storyboards.